\pdfoutput=1

\documentclass[11pt]{article}

\usepackage{algorithm} 
\usepackage{template/acl}
\usepackage{hyperref}
\usepackage{times}
\usepackage{latexsym}
\usepackage{multirow}
\usepackage{booktabs}
\usepackage{microtype}
\usepackage{multirow}
\usepackage{latexsym}
\usepackage{booktabs}
\usepackage{courier}
\usepackage{amsmath}
\usepackage{booktabs} 
\usepackage{siunitx} 
\usepackage{subfig}
\usepackage{placeins}
\usepackage{tikz}
\usepackage{float}
\usepackage{hyperref}

\usepackage{xcolor}
\usepackage{pgfplots}
\usepackage{tikz}

\usepackage[T1]{fontenc}

\usepackage[utf8]{inputenc}

\usepackage{microtype}

\usepackage{inconsolata}

\title{Lisbon Computational Linguists at SemEval-2024 Task 2: Using A Mistral-7B Model and Data Augmentation}

\author{Artur Guimarães \\
  INESC-ID and IST\\
  University of Lisbon \\
  Lisbon, Portugal \\
  \texttt{\small artur.guimas@gmail.com} \\ 
  \And
  Bruno Martins\\
  INESC-ID and IST\\
  University of Lisbon \\
  Lisbon, Portugal \\
  \texttt{\small bruno.g.martins@tecnico.ulisboa.pt} \\
  \And
  João Magalhães \\
  NOVA-LINCS\\
  NOVA University of Lisbon\\
  Lisbon, Portugal \\
  \texttt{\small jmag@fct.unl.pt} \\}

\begin{document}

\makeatletter
\renewcommand{\ALG@name}{Listing}
\makeatother

\maketitle
\begin{abstract}
    This paper describes our approach to the SemEval-2024 safe biomedical Natural Language Inference for Clinical Trials (NLI4CT) task, which concerns classifying statements about Clinical Trial Reports (CTRs). We explored the capabilities of Mistral-7B, a generalist open-source Large Language Model (LLM). We developed a prompt for the NLI4CT task, and fine-tuned a quantized version of the model using an augmented version of the training dataset. The experimental results show that this approach can produce notable results in terms of the macro F1-score, while having limitations in terms of faithfulness and consistency. All the developed code is publicly available on a GitHub repository\footnote{\href{https://github.com/araag2/SemEval2024-Task2}{https://github.com/araag2/SemEval2024-Task2}}. 
\end{abstract}

\section{Introduction}
Large Language Models (LLMs) currently achieve state-of-the-art performance on different Natural Language Processing (NLP) tasks, including in the assessment of textual entailment relations. However, these models are heavily susceptible to shortcut learning \cite{du2023shortcut}, factual inconsistency \cite{xie2023faithful}, and performance degradation when exposed to data from specialized domains, such as in the case of medical data.

Noting the aforementioned challenges, Task 2 at SemEval-2024 addressed a safe biomedical Natural Language Inference for Clinical Trials (NLI4CT) task~\cite{jullien-etal-2024-semeval}, which concerns classifying statements about Clinical Trial Reports (CTRs). NLI4CT investigated the accuracy, faithfulness, and consistency of the reasoning performed by LLMs in this particular medical task.
The goal of the task is to determine whether there is an entailment relation or a contradiction relation between CTRs and statements, making some type of claim about a single CTR or a pair of CTRs. 
Given the specific focus on assessing model faithfulness and consistency (i.e., the ability to make correct predictions for the correct reasons), the dataset associated to the task involved the systematic application of controlled interventions, either preserving or inverting the entailment relations originally generated by clinical domain experts. This way, the task investigated the robustness of NLI models in their representation of the semantic phenomena necessary for complex inference in clinical settings.

Our approach to the NLI4CT task involved the use of open-source LLMs, with good results in general purpose benchmarks\footnote{\href{https://huggingface.co/spaces/lmsys/chatbot-arena-leaderboard}{https://huggingface.co/spaces/lmsys/chatbot-arena-leaderboard}} and capable of following task instructions. We opted for \texttt{Mistral-7B-Instruct-v0.2}\footnote{\href{https://huggingface.co/mistralai/Mistral-7B-Instruct-v0.2}{https://huggingface.co/mistralai/Mistral-7B-Instruct-v0.2}} \cite{jiang2023mistral}, quantizing the model to 4-bits and simultaneously using Low-Rank Adaptation (LoRA) \cite{hu2021lora, dettmers2023qlora} to fine-tune the model to the NLI4CT task, using a slightly augmented version of
the training dataset that features a mixture of manually curated and synthetic statements.

Our overall best submission to the task achieved a \textbf{macro F1-score} of 0.80 (1st place on the leaderboard), a \textbf{consistency} score of 0.72 (15th), and a \textbf{faithfulness} score of 0.83 (11th). Our method excels in classification accuracy, but fails at being robust to perturbations on the statements, i.e. predicting the same label on contradictory examples and different labels on paraphrased examples. 

\section{Background}

The NLI4CT task concerns inferring if statements can be entailed by a given textual context, with each statement referring to one or two CTRs. These CTRs belong to a corpus consisting of 1000 different trials concerning breast cancer, extracted from the United States National Library of Medicine\footnote{\href{https://clinicaltrials.gov/}{https://clinicaltrials.gov/}}.
These trial reports are exclusively written in the English language and average 817 words in length.

CTRs are divided into four sections: \textbf{Eligibility Criteria}, describing a set of conditions to allow or exclude patients in the trial; \textbf{Interventions}, detailing all information about the conducted treatments; \textbf{Results}, outlining outcomes and experimental results gathered through the trial; and \textbf{Adverse Events}, reporting patient observations concerning symptoms and physiological signs. An instance of the NLI4CT dataset contains either one or two CTRs (i.e., cases denoted as single or comparison, respectively), a statement, a section marker, and a ground-truth label (i.e., entailment or contradiction). An example is shown next.

\begin{algorithm}
\caption{An instance from the NLI4CT dataset.}
\footnotesize 
\label{sec2:task_example}
        \texttt{Primary Trial:} \\ INTERVENTION 1: \\
            • Letrozole, Breast Enhancement, Safety. \\
            • Single arm of healthy postmenopausal women to have two breast MRI (baseline and post-treatment). Letrozole of 12.5 mg/day is given for three successive days just prior to the second MRI. \vspace{0.1cm} \\ \\
        \texttt{Secondary Trial:} \\ INTERVENTION 1: \\
             • FFDM Mammography Exam - LIP Algorithm \\
             • Screening or diagnostic Full Field Digital Mammography (FFDM) exam \\
             INTERVENTION 2: \\
             • FFDM Mammography Exam - SIP Algorithm. \\
             • The same 130 raw data images were externally reprocessed with the Siemens processing algorithm. \\ \\
        \texttt{Section:} Intervention \\
	\texttt{Statement:} The primary trial and the secondary trial both used MRI for their interventions. \\
        \texttt{Label:} Entailment.
\end{algorithm}

The dataset\footnote{\href{https://github.com/ai-systems/Task-2-SemEval-2024/blob/main/README.md}{https://github.com/ai-systems/Task-2-SemEval-2024/blob/main/README.md}} provided by the task organizers considered training, development, practice-test, and test splits (the last two without ground-truth labels during the competition), with a general statistical characterization provided in Table~\ref{sec2:task-data}. 

\begin{table}[t]
\centering
\resizebox{\columnwidth}{!}{
    \begin{tabular}{@{}lccc@{}}
        \toprule
        \textbf{Set} & \textbf{\# Samples} & \textbf{Single - Compari.} & \textbf{Entail. - Contr.} \\
        \midrule
        Training     & 1700  & 60.9\% - 39.1\% & 50\% - 50\% \\
        Development  & 200   & 70\% - 30\%     & 50\% - 50\% \\
        Pratice-test & 2142  & 71.2\% - 28.8\% & 34.1\% - 65.9\% \\
        Test         & 5500  & 46.4\% - 53.6\% & 33.5\% - 66.5\% \\
        \bottomrule
    \end{tabular}%
}
\caption{The NLI4CT task dataset.}
\label{sec2:task-data}
\end{table}

The first two splits, i.e. training and development, are similar to those used in the SemEval-2023 edition of the task~\cite{jullien-etal-2023-semeval}, based on the work by~\citet{jullien-etal-2023-nli4ct}. These are two balanced sets, with mostly unique CTR-statement associations (i.e., statements that are not rephrasing or contradicting other ones). On the other hand, this composition contrasts with the practice-test and test splits, that are both imbalanced and almost solely composed of statements featuring interventions (e.g., paraphrasing, contradicting, or appending text) over a small set of original statements (< 10\%), as show in Table \ref{sec2:alterations-test}. This distribution favours systems that focus on robustly classifying a small set of samples.     

\begin{table}[t]
\centering
\resizebox{\columnwidth}{!}{
    \begin{tabular}{@{}lcc@{}}
        \toprule
        \textbf{Set} & \textbf{\# Interventions} & \textbf{Preserving - Altering (Label)} \\
        \midrule
        Pratice-test & 1942 (90.7\%) & 82.7\% - 27.3\% \\
        Test & 5000 (90.9\%) & 82.7\% - 27.3\% \\
        \bottomrule
    \end{tabular}%
}
\caption{Interventions over statements on the test splits.}
\label{sec2:alterations-test}
\end{table}

\section{System Overview} \label{sec3}

We now describe our general approach to the SemEval-2024 NLI4CT task.

\subsection{Choice of LLM}

When deciding on how to build our NLI4CT system, we started by testing the zero-shot and few-shot capabilities of several open-source LLMs, before settling on the use of \texttt{Mistral-7B-Instruct-v0.2}. In addition to achieving good zero-shot results, this model also allowed us to process arbitrarily long input texts, which in this task is particularly relevant, since some CTRs can exceed 3000 tokens in length. 

\subsection{Model Prompting}
A great deal of attention is currently given to prompting techniques, as the successful use of an LLM can be severely impaired by suboptimal prompts, and also since instruction fine-tuning \cite{chowdhery2022palm, chung2022scaling} is dependant on the prompt quality. In order to address the task of choosing a good prompt, we started by creating a prompt template that we deemed as suitable for the task at hand, sub-dividing our prompt into distinct parts (pre-pended with ``\$'') that can latter be replaced with different textual realizations. The overall structure is illustrated next.

\begin{algorithm}[H]
\caption{Overall prompt structure.} 
 
 \footnotesize\label{sec3:section_parts}
    \texttt{\$task\_description} \vspace{0.05cm}  \\ 
    \texttt{\$ctr\_description} \vspace{0.15cm}  \\ ~ \\
    \texttt{Primary Trial:} \\
    \texttt{\$primary\_evidence} \vspace{0.07cm} \\ ~ \\
    \texttt{Secondary Trial:} \\
    \texttt{\$secondary\_evidence} \vspace{0.15cm}  \\ ~ \\
    \texttt{\$statement\_description} \vspace{0.05cm}  \\
    \texttt{\$statement} \vspace{0.1cm} \\ ~ \\
    \texttt{\$option\_description} 
\end{algorithm}

Four of the parts are \textbf{sample independent}: \texttt{\$task\_description} provides a general description for the natural language inference task between CTRs and statements; \texttt{\$ctr\_description} delineates the general contents of a CTR and its different sections; \texttt{\$statement\_description} conveys the nature of the \$statement; and lastly \texttt{\$option\_description} outlines the answers we expect from the model (e.g., an answer of YES or NO, depending on whether the CTR supports the statement). Conversely, \texttt{\$primary\_evidence}, \texttt{\$secondary\_evidence}, and \texttt{\$statement} are \textbf{sample dependent}, as these parts should be replaced by the primary CTR, the secondary CTR (if applicable), and the statement, respectively.

We created 5 base prompts (see Appendix~\ref{secA:base_prompts}) for each of the 4 sample independent parts, yielding 625 possible combinations for the general template. We evaluated all the combinations on the development set, and chose the prompt that yielded the top \textbf{macro F1-score}, which is shown in Listing 3.

\begin{algorithm}[t]
\caption{The best performing prompt.} 
\footnotesize
    <s>[INST]The objective is to examine semantic entailment relationships between individual sections of Clinical Trial Reports (CTRs) and statements articulated by clinical domain experts. CTRs elaborate on the procedures and findings of clinical trials, scrutinizing the effectiveness and safety of novel treatments. Each trial involves cohorts or arms exposed to distinct treatments or exhibiting diverse baseline characteristics. \\
    Comprehensive CTRs comprise four sections: (1) ELIGIBILITY CRITERIA delineating conditions for patient inclusion, (2) INTERVENTION particulars specifying type, dosage, frequency, and duration of treatments, (3) RESULTS summary encompassing participant statistics, outcome measures, units, and conclusions, and (4) ADVERSE EVENTS cataloging signs and symptoms observed. Statements posit claims regarding the information within these sections, either for a single CTR or in comparative analysis of two. To establish entailment, the statement's assertion should harmonize with clinical trial data, find substantiation in the CTR, and avoid contradiction with the provided descriptions. \\
    The following descriptions correspond to the information in one of the Clinical Trial Report (CTR) sections. \\
    
    Primary Trial: \\
    \textbf{\$primary\_evidence} \\
    
    Secondary Trial: \\ 
    \textbf{\$secondary\_evidence} \\
    
    Reflect upon the ensuing statement crafted by an expert in clinical trials. \\
    \textbf{\$statement} \\
    Respond with either YES or NO to indicate whether it is possible to determine the statement's validity based on the Clinical Trial Report (CTR) information, with the statement being supported by the CTR data and not contradicting the provided descriptions.[/INST] \textbf{Answer:}
\end{algorithm}

\subsection{Generating Answers} \label{sec3.3}

With the aforementioned template, we used the Python HuggingFace Transformers  library\footnote{\href{https://huggingface.co/docs/transformers/en/index}{https://huggingface.co/docs/transformers/en/index}} to generate answers with \texttt{Mistral-7B-Instruct-v0.2}, using as parameters \texttt{do\_sample=True}, \texttt{top\_k=5}, and \texttt{max\_new\_tokens=30}. We opted not to constrain the generation process, instead looking for sets of words, associated to each label, in the sequence of generated tokens. The words ``Yes'', ``yes'', and ``entailment'' were used for the entailment class, while the words ``No'', ``no'' and ``contradiction'' were used for the contradiction class. Preference was given to the first token in the sequence that belongs to either of the sets, and if none were found we label the instance as entailment.

\subsection{Data Augmentation} \label{sec3.4}

The NLI4CT dataset features 1700 training instances and 200 development instances, which is perhaps insufficient for fine-tuning an LLM in order to generalize to a testing split that is almost thrice as large. We decided to augment the available data, and created the 3 different training splits outlined in Table \ref{sec3:additional-data}.

\begin{table}[H]
\centering
\resizebox{\columnwidth}{!}{
    \begin{tabular}{@{}lccc@{}}
        \toprule
        \textbf{Set} & \textbf{\# Samples} & \textbf{Single - Compari.} & \textbf{Entail. - Contr.} \\
        \midrule
        Train\_Manual           & 2344  & 61.8\% - 38.2\%  & 50\% - 50\%     \\
        Train\_Manual-Synthetic & 3720  & 63.7\% - 36.3\%  & 50\% - 50\%     \\
        Train\_Full-Synthetic   & 11011 & 60.9\% - 39.1\%  & 46.3\% - 53.7\% \\
        \bottomrule
    \end{tabular}%
}
\caption{Results from task data augmentation.} 
\label{sec3:additional-data}
\end{table}

The three new sets were constructed as follows: 

\begin{itemize}
    \item \textbf{Train\_Manual:} Starting from the train split, we added queries created by using pre-existing samples with the entailment class, negating them using the Python negate library\footnote{\href{https://github.com/dmlls/negate}{https://github.com/dmlls/negate}} (i.e., to generate corresponding contradiction examples), and also manually paraphrasing the original instance (i.e., to generate different entailment samples). All 644 additional samples that were generated through this procedure were manually curated; 
    \item \textbf{Train\_Manual-Synthetic:} starting from the \textbf{Train\_Manual} dataset, we added 1376 new automatically generated instances to this set: half of the new instances were generated with the negate library, and the other half were generated by paraphrasing existing statements using the \texttt{Mistral-7B-Instruct-v0.2} model;
    \item \textbf{Train\_Full-Synthetic:} Starting from the train split, we added 9311 new samples, using the negate library on entailment instances, and the \texttt{Mistral-7B-Instruct-v0.2} model to paraphrase each original statement 5 times.
\end{itemize}

\subsection{Instruction Fine-tuning} \label{sec3.5}

Noting that \texttt{Mistral-7B-Instruct-v0.2} is a generalist instruction fine-tuned model, we sought to fine-tune this LLM to the NLI4CT task, using the aforementioned instructions. 
To improve the training efficiency and support very long sequences (i.e., up to 6000 tokens), we quantized the model to 4-bit representations of the parameters, and used LoRA~\cite{hu2021lora}. Model training used a supervised fine-tuning objective based on auto-regressive language modelling, completing the input instruction with the correct label for each instance (i.e., outputting either ``Yes'' or ``No'' after ``Answer:'' in the prompt). The implementation relied on the PEFT\footnote{\href{https://huggingface.co/docs/peft/en/index}{https://huggingface.co/docs/peft/en/index}} and TRL\footnote{\href{https://huggingface.co/docs/trl/en/index}{https://huggingface.co/docs/trl/en/index}} Python libraries.

\section{Experimental Setup}

Making official submissions to the task leaderboard required the participants to submit full runs of the test set, outputting a label for each of its instances. We obtained the labels for each instance by following the procedure described in Subsection \ref{sec3.3}.

The task uses the following evaluation measures: \textbf{macro F1-score}, i.e. the arithmetic mean of precision and recall, averaged over the two classes; \textbf{Faithfulness}, i.e. a measure created to assess the capacity of model to arrive at the correct prediction for the correct reason, calculated by measuring the ability of model to change its prediction label after semantically altering a statement; and \textbf{Consistency}, which completes faithfulness by measuring the ability of a model in outputting the same prediction for semantically equivalent statements\footnote{\href{https://github.com/ai-systems/Task-2-SemEval-2024/blob/main/evaluate.py}{https://github.com/ai-systems/Task-2-SemEval-2024/blob/main/evaluate.py}}. We evaluated our runs using the official metrics obtained from the leaderboard.

Following the training procedure described in Section \ref{sec3.5}, we tested different combinations of training data (as described in Section \ref{sec3.4}). The full set of hyper-parameters associated to our best run can be found in Appendix \ref{secA:hyper-parameters}. All the different runs used Python libraries and packages that can be found in our GitHub repository\footnote{\href{https://github.com/araag2/SemEval2024-Task2/blob/main/environment.yml}{https://github.com/araag2/SemEval2024-Task2/blob/main/environment.yml}}.

\section{Results and Discussion}
\label{sec5}
\begin{figure}[t]
\centering
    \begin{tikzpicture}
        \begin{axis}[
            width  = \columnwidth,
            height = 3.5cm,
            major x tick style = transparent,
            ybar=2*\pgflinewidth,
            bar width=8pt,
            ymajorgrids = true,
            ylabel = {Score},
            symbolic x coords={F1-Score, Faithfulness, Consistency},
            xtick = data,
            scaled y ticks = false,
            enlarge x limits=0.25,
            ymin=0,
            ymax=1,
            legend cell align=left,
            legend style={
                    scale=0.5,
                    at={(1.05, 0.85)},
                    anchor=south east,
                    column sep=1ex
            }
        ]
            \addplot[style={teal, fill=teal, mark=none}]
                coordinates {(F1-Score, 0.80) (Faithfulness, 0.83) (Consistency, 0.72)};
    
            \addplot[style={yellow, fill= yellow, mark=none}]
                 coordinates {(F1-Score, 0.80) (Faithfulness, 0.95) (Consistency, 0.81)};
    
            \addplot[style={gray, fill=gray,mark=none}]
                 coordinates {(F1-Score, 0.78) (Faithfulness, 0.95) (Consistency, 0.78)};
    
            \addplot[style={brown, fill=brown, mark=none}]
                 coordinates {(F1-Score, 0.78) (Faithfulness, 0.92) (Consistency, 0.77)};
    
            \legend{Ours, 1st, 2nd, 3rd}
        \end{axis}
    \end{tikzpicture}
\caption{Comparison of top submissions against our system, according to different evaluation metrics.}
\label{sec5:leaderboard}
\end{figure}
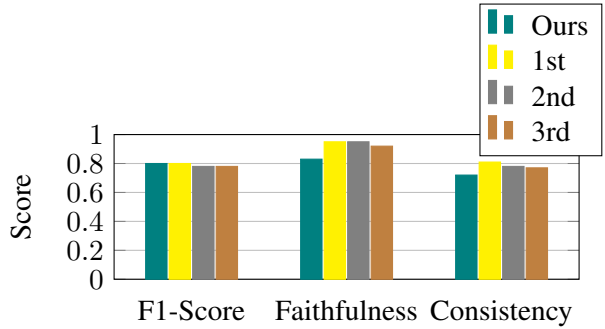

Table \ref{sec5:main-results} presents our most important results, showing the best result that we achieved with each training set. In turn, Figure \ref{sec5:leaderboard} compares our overall best run with the top three submissions, per metric. 

\begin{table}[H]
\centering
\resizebox{\columnwidth}{!}{
\begin{tabular}{@{}lccc@{}}
\toprule
\textbf{Trained Sets} & \textbf{F1-Score} & \textbf{Faithfulness} & \textbf{Consistency} \\
\midrule
None (Zero-Shot)          & 0.67 (3) & 0.61 (8) & 0.53 (8) \\
Train                     & 0.81 (2) & 0.72 (3) & 0.69 (2) \\
Train\_Manual             & \textbf{0.82 (9)} & 0.76 (9) & 0.71 (9) \\
 \textbf{Train\_Manual-Synthetic}   & 0.80 (1) & \textbf{0.83 (1)} & \textbf{0.72 (2)} \\
Train\_Full-Synthetic     & 0.78 (1) & 0.78 (0) & 0.71 (0) \\
\bottomrule
\end{tabular}%
}
\caption{Results on different training datasets.}
\label{sec5:main-results}
\end{table}

We tested the LLM without any training (i.e., zero-shot results), and fine-tuning with the base and augment datasets, all with our best instruction format. As expected, there is a significant difference in performance towards fine-tuned models. Overall, a mixture between manually curated samples and synthetically generated ones performed best (Train\_Manual-Synthetic, as described on Section \ref{sec5:main-results}), outperforming the best run that did not use any data augmentation. If more instances could have been manually curated, specifically targeting adversarial re-writes of the same statements, we hypothesize that results could be improved further. Even though \textbf{Train\_Full-Synthetic} corresponds to the largest training set (i.e., featuring 11011 samples), the lack of quality in the automatically generated statements potentially impaired the \textbf{F1-score} while also limiting \textbf{consistency} and \textbf{faithfulness}. 

The run trained with \textbf{Train\_Manual-Synthetic} corresponds to our best overall result. When compared to the top submissions, we can see that our F1-score corresponds to a tie with another system in the 1st place of the leaderboard. However, results are much worse in the other two metrics, with significant differences between the top systems (i.e., with scores of 0.95 in faithfulness and 0.81 in consistency) and our submission.

In the post-task phase of the competition, ground-truth labels for all examples were released, specifying which type of interventions were made in each instance. Therefore, we are now able to analyse our system's errors (see Table \ref{sec5:error-analysis}), to support a discussion on the main short-comings of our work. 

\begin{table}[H]
\centering
\resizebox{\columnwidth}{!}{
\begin{tabular}{@{}lc@{}}
\toprule
\textbf{Type of Error} & \textbf{\# Occurrences / \# Total Samples}\\
\midrule
Base Statement Errors & 99 / 500 (19.8\%) \\
Intervention Errors   & 1328 / 5000 (26.7\%) \\
Total Errors          & 1427 / 5500 (25.9\%) \\
\midrule
Label Preserving Intervention Errors & 1177 / 1328 (88.6\%) \\
Label Altering Intervention Errors   & 151 / 1328  (11.4\%) \\
\midrule
Paraphrasing Errors            & 344 / 1500 (22.9\%) \\
Text Appending Errors          & 609 / 1500 (40.6\%) \\
Contradicting Errors           & 293 / 1500 (19.5\%) \\
Numerical Paraphrasing Errors  & 58 / 224 (25.9\%) \\
Numerical Contradicting Errors & 24 / 276 (8.7\%) \\
\bottomrule
\end{tabular}%
}
\caption{Error analysis for our best overall run, categorizing errors by intervention types.}
\label{sec5:error-analysis}
\end{table}

Comparing all errors across the different instance types, the average error rate is much higher on intervention errors (26.7\%) against base statement errors (19.8\%), which is to be expected as our training sets had fewer examples of this type. Specifically, we can see that label preserving interventions (88.6\%) have a high percentage of errors.
Our system can identify instances which suffered contradictory interventions with an error rate of 19.5\% for textual changes, and 8.7\% for numerical changes. Instances that were perturbed with paraphrasing cause an error rate of 22.9\%, while numerical paraphrasing errors correspond to 25.9\%.
At the worst end we have the samples with text appended to the end, which causes an error rate of 40.6\%. Note that we did did not explicitly augment the training instances by appending text to the existing statements, and the absence of examples like this was very costly in terms of the final results.

\section{Conclusions}

Adapting evaluation methodologies to better inform the safe deployment of LLMs in critical domains is an urgent necessity. The NLI4CT task at SemEval-2024 addressed this specific concern, and through our participation we improved our understanding on how LLMs can be fine-tuned to encompass robust results on clinical natural language inference. Overall, our results show that the simple fine-tuning of an open-source LLM to this specific task can achieve notable results in terms of the macro-averaged F1, although with limitations in terms of faithfulness and consistency. Augmenting the data with high-quality curated examples can improve result quality, although augmenting the training set with synthetic examples requires careful quality control.

For future work we would like to explore the following ideas:

\begin{itemize}
    \item Test our general approach with different models, specifically considering models fine-tuned in the medical domain (e.g., models like \texttt{qCammel-70-x}\footnote{\href{https://huggingface.co/augtoma/qCammel-70-x}{https://huggingface.co/augtoma/qCammel-70-x}} or \texttt{BioMistral}\footnote{\href{https://huggingface.co/BioMistral}{https://huggingface.co/BioMistral}});
    \item Refining the considered prompt through recently-proposed prompt optimization methods \cite{wen2023hard,guo2023connecting}, instead of relying on manually curated prompts;
    \item Incorporating additional training data, e.g. by generating a more diverse set of instances from the CTR data made available in the context of other shared tasks (e.g., the CTR data from the Text Retrieval Conference (TREC) clinical trials track\footnote{\href{https://www.trec-cds.org/}{https://www.trec-cds.org/}});
    \item Carefully curating a new training set, with a focus on statement interventions rather than quantity of base statements, in order to better guide the model into understanding the nuances of textual and numerical paraphrasing/contradiction.
\end{itemize}

\section*{Acknowledgments}

This research was supported by the Portuguese Recovery and Resilience Plan through project C645008882-00000055 (i.e., the Center For Responsible AI), and also by Fundação para a Ciência e Tecnologia (FCT), through the project with reference UIDB/50021/2020 (DOI:10.54499/UIDB/50021/2020).

\bibliography{main_bib}

\appendix

\section{Appendix}
\label{sec:appendix}

We now present additional details about the prompt considered for instructing the \texttt{Mistral-7B-Instruct-v0.2} model, and about the hyper-parameters considered for model fine-tuning.

\subsection{Base Descriptions For Each Prompt Part} \label{secA:base_prompts}

This section presents the five different alternatives that were considered for the different parts of the \texttt{Mistral-7B-Instruct-v0.2} prompt.

\subsubsection{Task Description Part}\scriptsize

1 : Consider the task of determining semantic entailment relations between individual sections of Clinical Trial Reports (CTRs) and statements made by clinical domain experts. Note that CTRs outline the methodology and findings of a clinical trial, which are conducted to assess the effectiveness and safety of new treatments. Each trial involves 1-2 patient groups, called cohorts or arms, and these groups may receive different treatments, or have different baseline characteristics. The complete CTRs contain 4 sections, corresponding to (1) a list of the ELIGIBILITY CRITERIA corresponding to the conditions for patients to be allowed to take part in the clinical trial, (2) a description for the INTERVENTION that specifies the type, dosage, frequency, and duration of treatments being studied, (3) a summary of the RESULTS, detailing aspects such as the number of participants in the trial, the outcome measures, the units, and the conclusions, and (4) a list of ADVERSE EVENTS corresponding to signs and symptoms observed in patients during the clinical trial. In turn, the statements are sentences that make some type of claim about the information contained in one of the aforementioned sections, either considering a single CTR or comparing two CTRs. In order for the entailment relationship to be established, the claim in the statement should be related to the clinical trial information, it should be supported by the CTR, and it must not contradict the provided descriptions. \\ ~ \\  \noindent

\noindent 2 : You are tasked with determining support relationships between individual sections of Clinical Trial Reports (CTRs) and clinical statements. CTRs detail the methodology and findings of clinical trials, assessing effectiveness and safety of new treatments. CTRs consist of 4 sections: (1) ELIGIBILITY CRITERIA listing conditions for patient participation, (2) INTERVENTION description specifying type, dosage, frequency, and duration of treatments, (3) RESULTS summary detailing participants, outcome measures, units, and conclusions, and (4) ADVERSE EVENTS listing signs and symptoms observed. Statements make claims about information in these sections, either for a single CTR or comparing two. \\ ~ \\  

\noindent 3 : Evaluate the semantic entailment between individual sections of Clinical Trial Reports (CTRs) and statements issued by clinical domain experts. CTRs expound on the methodology and outcomes of clinical trials, appraising the efficacy and safety of new treatments. The statements, on the other hand, assert claims about the information within specific sections of CTRs, for a single CTR or comparative analysis of two. For entailment validation, the statement's claim should align with clinical trial information, find support in the CTR, and refrain from contradicting provided descriptions. \\ ~ \\ 

\noindent 4 : The objective is to examine semantic entailment relationships between individual sections of Clinical Trial Reports (CTRs) and statements articulated by clinical domain experts. CTRs elaborate on the procedures and findings of clinical trials, scrutinizing the effectiveness and safety of novel treatments. Each trial involves cohorts or arms exposed to distinct treatments or exhibiting diverse baseline characteristics. Comprehensive CTRs comprise four sections: (1) ELIGIBILITY CRITERIA delineating conditions for patient inclusion, (2) INTERVENTION particulars specifying type, dosage, frequency, and duration of treatments, (3) RESULTS summary encompassing participant statistics, outcome measures, units, and conclusions, and (4) ADVERSE EVENTS cataloging signs and symptoms observed. Statements posit claims regarding the information within these sections, either for a single CTR or in comparative analysis of two. To establish entailment, the statement's assertion should harmonize with clinical trial data, find substantiation in the CTR, and avoid contradiction with the provided descriptions. \\ ~ \\

\noindent 5 : Consider the problem of assessing semantic entailment connections between distinct sections of Clinical Trial Reports (CTRs) and statements put forth by clinical domain experts. To establish entailment, the statement's assertion should be supported from the CTR, not contradicting the provided descriptions. In brief, CTRs elucidate the procedures and findings of clinical trials, evaluating the efficacy and safety of emerging treatments. Complete CTRs encompass four sections: (1) ELIGIBILITY CRITERIA specifying conditions for patient inclusion, (2) INTERVENTION details on the type, dosage, frequency, and duration of treatments, (3) RESULTS summarizing the participant statistics, outcome measures, units, and conclusions, and (4) ADVERSE EVENTS listing observed signs and symptoms. Statements advance claims about the information within these sections, either for a single CTR or in a comparative analysis of two CTRs.

\subsubsection{CTR Description Part}\scriptsize
1 : The following descriptions correspond to the information in one of the Clinical Trial Report (CTR) sections. \\ ~\\

\noindent 2 : The provided descriptions coincide with the content in a specific section of Clinical Trial Reports (CTRs), detailing relevant information to the trial. \\ ~ \\

\noindent 3 : The provided descriptions correspond to the content found in one of the four standard clinical trial report sections. \\ ~\\ 

\noindent 4 : The provided descriptions pertain to the contents found within one of the sections of Clinical Trial Reports (CTRs). \\ ~\\

\noindent5 : The descriptions that follow correspond to the information contained in one of the standard sections of the clinical trial reports. 

\subsubsection{Statement Description Part}
1 : Consider also the following statement generated by a clinical domain expert, a clinical trial organizer, or a medical researcher. \\ ~ \\
\noindent 2 : Contemplate the ensuing statement formulated by a clinical expert or researcher. \\ ~ \\
\noindent 3 : Review the subsequent statement provided by an expert in clinical trials, attending to the medical terminology and carefully addressing any ambiguities. \\ ~ \\
\noindent 4 : Deliberate upon the subsequent statement formulated by an healthcare practitioner, a coordinator of clinical trials, or a medical researcher. \\ ~ \\
\noindent 5 : Reflect upon the ensuing statement crafted by an expert in clinical trials.

\subsubsection{Option Description Part}
1 : Answer YES or NO to the question of whether one can conclude the validity of the statement with basis on the clinical trial report information. \\ ~\\ 

\noindent 2 : Indicate with either YES or NO whether it is possible to determine the validity of the statement based on the Clinical Trial Report (CTR) descriptions. An answer of YES means that the statement is supported by the CTR descriptions, not contradicting the provided information. \\ ~\\

\noindent 3 : Provide a YES or NO response indicating if it's possible to assess the statement's validity based on the information presented in the clinical trial report descriptions. Do this by interpreting the medical terminology and the context in both the report and the statement, carefully addressing any ambiguities or gaps in the provided information. \\ ~\\

\noindent 4 : Respond with either YES or NO to indicate whether it is possible to determine the statement's validity based on the Clinical Trial Report (CTR) information, with the statement being supported by the CTR data and not contradicting the provided descriptions. \\ ~ \\

\noindent 5 : Indicate with a YES or NO response whether it is possible to assess the statement's validity based on the clinical trial report data.

\subsection{Full List of Hyper-Parameters} \label{secA:hyper-parameters} 

\normalsize

\noindent
The full list of hyper-parameters considered for model fine-tuning can be seen in the source-code in our GitHub repository\footnote{\href{https://github.com/araag2/SemEval2024-Task2/blob/main/finetune_Mistral.py}{https://github.com/araag2/SemEval2024-Task2/blob/main/finetune\_Mistral.py}}.

~\\
\noindent
The chosen parameters concerning model quantization options are as follows.

~\\
\noindent
\texttt{load\_in\_4bit = True} \\
\texttt{bnb\_4bit\_quant\_type = "nf4"} \\
\texttt{bnb\_4bit\_compute\_dtype = torch.bfloat16} \\
\texttt{bnb\_4bit\_use\_double\_quant = False} \\

~\\
\noindent
The parameters concerning the use of Low-Rank Adaptation (LoRA) are as follows.

~\\
\noindent
\texttt{lora\_r = 64} \\
\texttt{lora\_dropout = 0.1} \\
\texttt{lora\_alpha = 16} \\
\texttt{bias = "none"} \\

\noindent
Finally, the general model training hyper-parameters are as follows.

~\\
\noindent
\texttt{train\_epochs = 5} \\
\texttt{batch\_size = 2} \\
\texttt{gradient\_accumulation\_steps = 4} \\
\texttt{learning\_rate = 2e-5} \\
\texttt{pooling = "mean"} \\
\texttt{max\_sequence\_length = 6000} \\

\end{document}